\documentclass{article}

\usepackage[preprint]{neurips_2021}
\pdfoutput=1

\usepackage[utf8]{inputenc} 
\usepackage[T1]{fontenc}    
\usepackage{hyperref}       
\usepackage{url}            
\usepackage{booktabs}       
\usepackage{amsfonts}       
\usepackage{nicefrac}       
\usepackage{microtype}      
\usepackage{xcolor}         

\usepackage{graphicx}
\usepackage{microtype}
\usepackage{wrapfig}
\usepackage{graphicx}
\usepackage{booktabs} 
\usepackage{algorithm}
\usepackage{algpseudocode}
\usepackage{amsmath}
\usepackage{amssymb}
\usepackage{adjustbox}
\usepackage{capt-of}

\usepackage{hyperref}
\usepackage{amsmath}
\usepackage[]{authblk} %
\newtheorem{definition}{Definition}
\newtheorem{theorem}{Theorem}
\DeclareMathOperator{\MVR}{MVR}
\DeclareMathOperator{\MV}{MV}
\DeclareMathOperator{\MM}{MM}
\DeclareMathOperator{\BL}{BL}
\title{MOFA: Modular Factorial Design for Hyperparameter Optimization}

\author[1\thanks{Work done as an internship at Huawei Noah's Ark Lab}]{\textbf{Bo Xiong} }
\author[2]{\textbf{Yimin Huang}}
\author[2]{\textbf{Hanrong Ye}}
\author[1,3]{\textbf{Steffen Staab}}
\author[2]{\textbf{Zhenguo Li}}
\affil[1]{University of Stuttgart, Germany}
\affil[2]{Huawei Noah's Ark Lab, China}
\affil[3]{University of Southampton, United Kingdom}

\begin{document}
\maketitle

\begin{abstract}
This paper presents a novel and lightweight hyperparameter optimization (HPO) method, MOdular FActorial Design (MOFA). MOFA pursues several rounds of HPO, where each round alternates between \emph{exploration} of hyperparameter space by \emph{factorial design} and \emph{exploitation} of evaluation results by \emph{factorial analysis}.
Each round first \emph{explores} the configuration space by constructing a low-discrepancy set of hyperparameters that cover this space well while de-correlating hyperparameters, and then \emph{exploits} evaluation results through factorial analysis that determines which hyperparameters should be further explored and which should become fixed in the next round. 
We prove that the inference of MOFA achieves higher confidence than other sampling schemes. Each individual round is highly parallelizable and hence offers major improvements of efficiency compared to model-based methods. Empirical results show that MOFA achieves better effectiveness and efficiency compared with state-of-the-art methods. 
\end{abstract}

\section{Introduction}
Modern machine learning techniques have achieved promising results in various areas ~\cite{szegedy2016rethinking,redmon2016you,vaswani2017attention,xiong2020learning}. However, the performance of these models highly depends on the configurations of their hyperparameters \cite{feurer2019hyperparameter}. 
For example, the performance of deep neural networks may fluctuate dramatically under different neural architectures~\cite{liu2018darts,liu2018progressive}. Different data augmentation policies can lead to different experimental results for an image recognition task ~\cite{cubuk2019autoaugment}. 
Heuristic tuning by using expert's experiences is a possible solution but only works on simple settings. 
\par
To avoid dependence on experts' experiences, automated hyperparameter optimization (HPO) \cite{yao2018taking,feurer2015efficient} searches for optimal hyperparameters. Currently, there are two lines of works on HPO. 
(1) \emph{Model-based methods} such as Bayesian Optimization (BO) \cite{mockus1978application} optimize hyperparameters by learning a surrogate model (e.g. Gaussian process). While there are some more sophisticated approaches such as SMAC \cite{hutter2011sequential}, BOHB \cite{falkner2018bohb} and Reinforcement Learning (RL) \cite{zoph2016neural}, all these methods highly depend on the model parametrization and cannot be fully parallelized as their core strategies iteratively pursue improvements. 
(2) \emph{Model-free methods} (e.g. Random Search) do not depend on any parametric model and can run in fully parallel fashion, as the different hyperparameter configurations are run and evaluated individually \cite{yu2020hyper}. Nevertheless, the sample efficiency of Random Search is not on par with model-based methods to achieve a globally optimal result. 
Besides, some works studied how to automatically change the search space during optimization \cite{ha2019bayesian} and explored identifying hyperparameter importance \cite{hutter2014efficient, van2018hyperparameter}, but these works still rely on the model parametrization. 
\par
Factorial designs \cite{wu2011experiments}, which deliberate how to adjust parameters and exploit corresponding experimental responses, can be used to improve the sample efficiency while avoiding model parametrization. Some HPO methods have begun to make use of factorial designs, such as Latin Hypercubes \cite{brockhoff2015impact,konen2011tuned} and Orthogonal Arrays (OAs) \cite{zhang2019deep}. 
So far, however, factorial designs for HPO have only been applied to explore the hyperparameter space without exploiting possible feedback from the evaluation results returned by running various hyperparameter configurations. 
\par
This paper present MOdular FActorial Design (MOFA), a multi-module process that combines factorial design with factorial analysis, which achieves high quality of optimization results while allowing for high efficiency through excellent parallelizability. The main idea of MOFA is to improve \emph{exploration} of hyperparameter sampling with \emph{factorial designs} and improve \emph{exploitation} of evaluation results with \emph{factorial analysis} by identifying the hyperparameter importance and reducing the search space. 
In each round, MOFA first \emph{explores} the configuration space by constructing a low-discrepancy set of hyperparameters that cover this space well while de-correlating hyperparameters and then \emph{exploits} evaluation results through factorial analysis that determines which hyperparameters should be further explored and which should become fixed in the next round.
Specifically, MOFA runs through four modules in each round: Firstly, an Orthogonal Latin Hypercube (OLH)-based sampler ensuring both univariate projection uniformity (low-discrepancy) and orthogonality, which explores the hyperparameter space more efficiently without correlating the hyperparemeters (improving \textit{exploration}). Secondly, a highly parallelized evaluator. Thirdly, a transformer collapsing the OLH performance table into an Orthogonal Array (OA). Finally, factorial analysis narrows down the search space and selects hyperparameters that are most promising for iterative optimization. 

To summarize, the main contributions of our paper are: 
\begin{itemize}
\item We propose MOFA, a novel and lightweight HPO method that shares the advantages of being model-free, parallelizable and sample efficient. 
\item To our best knowledge, we are the first to exploit factorial design with factorial analysis on the setting of HPO. 
\item We provide necessary theoretical analysis to show how MOFA controls the worst-case response and its inference reliability. 
\item Empirical results show that MOFA clearly improves effectiveness and efficiency of HPO compared to state-of-the-art methods. 
\end{itemize}

\section{Preliminaries}
\begin{figure*}[t!]
\begin{center}
\includegraphics[width=0.9\columnwidth]{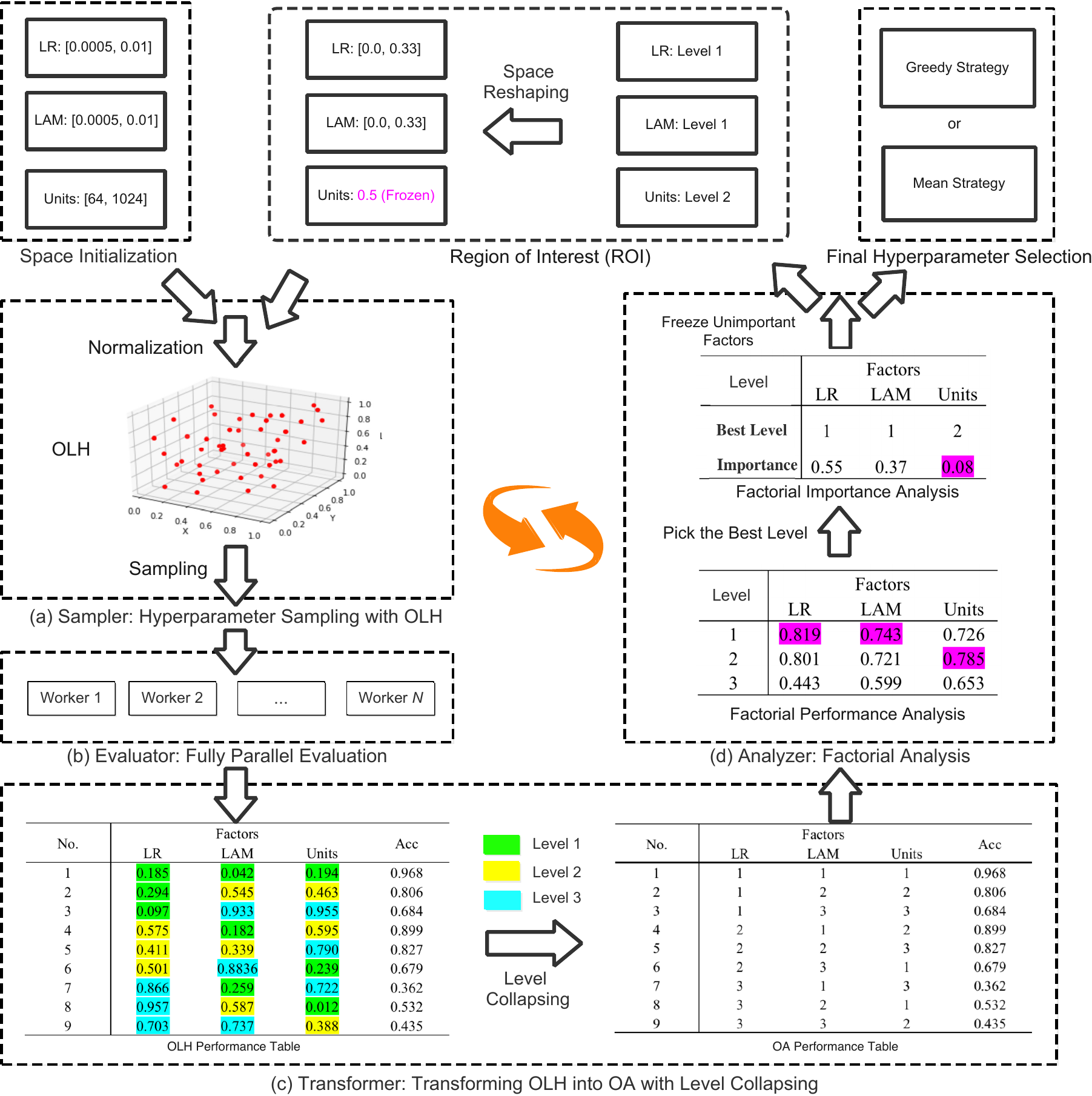}
 \vspace*{-2mm}
\caption{The overview of MOFA. MOFA consists of four modules. (a) an OLH-based sampler for hyperparameter sampling; (b) a paralleled evaluator; (c) a transformer collapsing OLH performance table into an OA. (d) an analyzer narrowing down the search space.}
\label{fig:bigmap}
\label{fig:bigmapa}
\label{fig:bigmapb}
\label{fig:bigmapc}
\end{center}
\vskip -0.07in
\end{figure*}

\textbf{Latin Hypercubes. } A Latin Hypercube \cite{mckay1979comparison} is an $N\times d$ table for an $N$-run experiment with $d$ factors, which is based on the Latin Square in which there is only one point in each row and column of a gridded space. A Latin Hypercube is the generalization of the Latin Square to an arbitrary number of dimensions, whereby each sample is the only one in each axis-aligned hyper-plane containing it. Fig.~\ref{fig:bigmap}c (left) shows an example of a Latin Hypercube with three factors. A Latin Hypercube has the property of univariate or one-dimensional projection uniformity which means that by projecting an $l$-point design on to any factor we will get $l$ different levels for that factor \cite{tang1993orthogonal}. This property drastically reduces the number of configurations necessary to achieve a reasonably accurate result.   


\textbf{Orthogonal Arrays.}  An OA is an $N\times d$ table, and each factor has $l$ levels. For any $t$ columns, the different level configurations appear with equal frequency ($t$-dimensional orthogonality). The number $t$ is called the strength. A Latin Hypercube is a special OA of strength one. For example, Fig.~\ref{fig:bigmapc}c (right) is an $OA(9,3,3,2)$, where the number of runs is nine, the number of factors is three, the number of levels is three and the strength is two. In any selected two columns, all possible configurations $(1,1), (1,2), (1,3), (2,1), (2,2), (2,3), (3,1), (3,2), (3,3)$ appear (\emph{complete}) and appear the same number of times (\emph{balanced}). This number is called the \emph{index} ($\lambda$). In OA design, the size of OA should meet the restriction $N=\lambda l^t$, where $l$ should be a prime number. 
The orthogonality in OA ensures that each factor's main effect can be
determined unaffected by interaction with other factors.
 
\textbf{Orthogonal Latin Hypercubes.} A randomly generated Latin Hypercube may be quite structured: the design may not have good univariate projection uniformity or the different factors might be highly correlated \cite{joseph2008orthogonal}. Several criteria such as maximin distance \cite{franco2008exploratory} and minimum correlation \cite{tang1998selecting} have been proposed to address these issues. Since OA ensures that the factor's main effect should not be influenced by its interaction effect with others, we use OA to construct Orthogonal Latin Hypercubes (OLH) \cite{tang1993orthogonal}). Beyond non-orthogonal Latin Hypercubes, OLH ensures both univariate projection uniformity and $t$-dimensional orthogonality, making it more suitable for factorial analysis. 

\section{Modular Factorial Design for Hyperparameter Optimization}
Fig.~\ref{fig:bigmap} shows an overview of MOFA. MOFA starts with a space initialization. Then, MOFA runs through four modules in each iteration. (1) \emph{Sampler} (Fig.~\ref{fig:bigmap}a): we construct an OLH ensuring both univariate projection uniformity and orthogonality to sample hyperparameter configurations. (2) \emph{Evaluator} (Fig.~\ref{fig:bigmap}b): the sampled hyperparameter configurations are evaluated in parallel. (3) \emph{Transformer} (Fig.~\ref{fig:bigmapc}c): an OLH performance table is built based on the evaluated results and collapsed into an OA performance table. (4) \emph{Analyzer} (Fig.~\ref{fig:bigmap}d): factorial performance analysis and factorial importance analysis are conducted to narrow down the search space and select influential hyperparameters. 

\subsection{Sampler: Hyperparameter Sampling with OLH}
\label{sec: sample}
 
In the sampler (Fig.~\ref{fig:bigmap}a), we first normalize the search space for each (discrete or continuous) hyperparameter into $[0,1]$ so that each hyperparameter is located in the same sampling space. Secondly, we build an OLH that ensures both one-dimensional projection uniformity and $t$-dimensional orthogonality to sample hyperparameter configurations. The one-dimensional projection uniformity makes it explore the search space more efficiently, while orthogonality reduces the interaction between hyperparameters to evaluate their main effects, making the validation results being more suitable for factorial analysis. 
For a specific HPO task, an OLH with a set of factors and a number of runs must be specified. Each factor corresponds to a hyperparameter. The number of runs should be chosen to meet the restrictions of OA design ($N=\lambda l^t$), where $l$ is the number of level, the index $\lambda$ describes the number of repeats of each level for each factor, $t$ is the strength that describes how many hyperparameters might simultaneously interact. MOFA offers to balance between a low polynomial and high expressiveness, e.g. a higher strength $t$ provides better decorrelation while the restriction of OA would be stricter, and vice versa.  Also, MOFA provides to balance between \emph{exploration} and  \emph{exploitation}, e.g. a larger index $\lambda$ makes MOFA \emph{explore} more hyperparameter configurations but consume more budget in a single round, and vice versa. One might also consider using more flexible designs such as near-orthogonal Latin Hypercube \cite{cioppa2007efficient} to avoid these restrictions, while more flexible design might suffer from some loss of performance. These interesting \emph{trade-off} problems are left for our future research. 

\subsection{Evaluator: Parallel Evaluation}
\label{sec: evaluator}

Different from model-based methods that learn a parametric model and update it based on previous experiences, the hyperparameter configurations sampled by OLH can be evaluated in parallel. In practice, we evaluate different hyperparameter configurations asynchronously on multiple workers. We analyze the parallelization in Sec.~\ref{sec: discussions}. 

\subsection{Transformer: Transforming OLH into OA}
\label{sec: transformer}

After evaluating the sampled hyperparameter configurations, we build an OLH performance table to store all the hyperparameter settings and evaluation results. By default, as depicted in 
Fig.~\ref{fig:bigmap}c (left), there are only continuous factor levels.
To allow for factorial analysis that only supports discrete levels, we collapse the continuous levels in each column of OLH into $L$ discrete ordered levels (highlighted with colors). Since the OLH meets orthogonality, the collapsed OLH will be an OA of the same size (corresponding to $N$ runs).

\subsection{Analyzer: Factorial Analysis}
\label{sec: analyzer}

\textbf{Factorial Performance Analysis:} Based on the OA, we first calculate the \emph{marginal mean} ($\MM$) performance for each factor $F^i$ at level $l$ by
$\MM(F^i_l)=\frac{L}{N}{\sum_{k\in [1...\frac{N}{L}]}{Y(F^i_{lk})}},$
where $Y(F^i_{lk})$ is the response for $F^i$ at level $l$ on its $k_{th}$ row. Since OA satisfies orthogonality, the $\MM$ can be seen as an approximation of the overall performance of each level for each factor, even when some factors are highly correlated (e.g. the \textit{learning rate} and the \textit{learning rate decay}). Then, we pick the level with the best $\MM$ performance of each factor for subsequent search with 
$
\BL(F^i)=\mbox{argmax}_{l\in [1...L]} \MM(F^i_l).
$
For example, in Fig.~\ref{fig:bigmap}d, the best $\MM$ performance of factor \textit{LR} is $0.819$, so the corresponding level $1$ is selected as the best level of \textit{LR}. Similarly, level $1$ and level $2$ are selected as the best levels of factor \textit{LAM} and \textit{Units}, respectively.

\textbf{Factorial Importance Analysis: } To narrow down the search space, we analyze the importance of each factor. We use the \emph{marginal variance ratio} 
$\MVR({F^i})=\frac{\MV ({F^i})}{\sum_{i=1}^{d} \MV ({F^i})},$ to measure the relative importance of each factor, 
where $\MV({F^i})={\mathbb{E}({\MM(F^i)} - \mathbb{E}({\MM(F^i}})))^2$ is the \emph{marginal variance} of factor $F^i$, $d$ is the number of factors. Since OA satisfies orthogonality, the $\MV({F^i})$ reflects the stability of results when varying factor $F^i$. Therefore, the $\MVR$ reflect its relative stability among all factors, which can be used to measure the relative importance of the factor. Specifically, if results produced by varying a factor are not stable, it indicates that the factor needs to be explored more. Conversely, if a factor is stable enough (the $\MVR$ is less than a specified threshold $\beta$), it does not need to be explored. In this case, we directly freeze it to be the current best level (median of the current search space). For example, in Fig.~\ref{fig:bigmap}d, the importance ($\MVR$) of factor \textit{LR}, \textit{LAM} and \textit{Units} in the current best level are $0.55$, $0.37$ and $0.08$ respectively, and the importance of \textit{Units} is less than the specified threshold $0.1$, so we freeze \textit{Units} to be $0.5$ (the median of search space $[0,1]$).




\subsection{Final Hyperparameter Selection}
\label{sec: final}
The termination condition can be designed case-by-case, e.g.\ defined by a a maximum budgets or stop if all the hyperparameters are freezed. Once the iteration ends, two strategies are available to select the final hyperparameter configuration. (1) \emph{Greedy Strategy}: we choose the configuration with the best performance among all of the evaluated experiments as the final hyperparameter setting. (2) \emph{Mean Strategy}: we do a factorial performance analysis for the hyperparameters that have not yet been determined and use the median level of the search space with the largest marginal mean performance as the final configuration. We combine these two strategies by picking the best hyperparameter configuration among these two strategies. 






\section{Theoretical Analysis}
\label{sec: theory}

In this section, the theoretical properties of MOFA are analyzed with regard to two aspects. 
First, we study how MOFA controls the worst-case response. Second, we prove that the inference of MOFA is more reliable. Precisely, if we make a hypothesis test for the inference of MOFA, it will have a higher confidence level.

\subsection{Worst-case Response}
For factorial analysis (Fig.~\ref{fig:bigmap}d), MOFA aims to compare different levels and choose the best one. The mean statistics $\bar Y_i=\sum_{X_j\in\text{level}~i} f(X_j)$ is used to estimate the effect of the $i$-th level, $\mathbb E_{X\in\text{level}~i} f(X)$, where $f(X)$ is the response of the factor vector $X$. Then, according to these estimates, the level with the highest mean statistics is picked as the best level. Although the estimator will approach the effect asymptotically, it can be poor when the dataset $P=(X_1,X_2,\ldots,X_N$) is small. In this case, one can use the Koksma-Hlawka inequality \cite{koksma1942een,hlawka1971discrepancy,aistleitner2015functions} to control the worst case. The Koksma-Hlawka inequality is stated as Eq.(\ref{eq:wcr}).
\begin{equation}\label{eq:wcr}
    |\mathbb E f(X)-\bar Y|\leq V(f)\cdot D^*(P)
\end{equation}
where $V(f)$ is the variation in the sense of Hardy and Krause \cite{pausinger2015koksma} and $D^*$ is the star discrepancy \cite{drmota2006sequences}. The variation $V(f)$ is a constant in our situation. Hence, reducing the star discrepancy of $P$ is a natural way to control the worst-case response.

\begin{definition}
The star discrepancy is defined as follows,
\begin{equation}
    D^*(P)=\sup_{(b_1,b_2\ldots,b_d)\in [0,1)^d}\left|{\frac{A(B;P)}{N}}-\lambda_{d}(B)\right|,
\end{equation}
where $\lambda_d$ is the $d$-dimensional Lebesgue measure, $A(B;P)$ is the number of points in $P$ that fall into $B$, and $B=\{(z_1,z_2\ldots,z_d)\in\mathbb R^d~|~0\leq z_j < b_j~\text{for}~j=1,2,\ldots,d\}.$
\end{definition}
Roughly speaking, the discrepancy of $P$ is low if the proportion of points in $P$ falling into an arbitrary set $B$ is close to proportional to the measure of $B$.

As mentioned before, MOFA has univariate projection uniformity. It is exactly the reason why we can control the worst case.
Notice that the one-dimensional projections of samples obtained from MOFA are much more evenly distributed than that of random sampling. For intuitive explanation, let $d=1$, we can see that the star discrepancy of MOFA with size $N$ is at most $1/N$ while the star discrepancy of random samplings of the same size has the order of $1/\sqrt{N}$ \cite{doerr2014lower}.

\subsection{Hypothesis Test}
For the MOFA procedure, we compare the mean performance of levels, and choose the level with highest response. To make this inference convincing, it is necessary to show that this level is statistically significantly better than other levels. Thus, a hypothesis test as follows is introduced to analyse this problem, 
\begin{align}\label{hypo}
    H_0:~&\mu_1\geq\mu_2,~\sigma_1 ~\text{and}~ \sigma_2~\text{are unknown, but cannot be}\nonumber\\
    &\text{assumed to be equal;}\\
    H_a:~&\mu_1<\mu_2, \sigma_1 ~\text{and}~ \sigma_2~\text{do not change,}\nonumber 
\end{align}
where $\mu_i$ and $\sigma_i$ is mean and standard deviation of the effect of the $i$-th level.
Let $\bar y_i$ and $s_i$ be the estimators of $\mu_i$ and $\sigma_i$, respectively.
Without loss of generality, assume $\bar y_1<\bar y_2$ and the inference is that the second level is better than the first level. Then, the $p$-value of the hypothesis test (\ref{hypo}) is used to measure the confidence of this inference. To show the advantage of MOFA, the following theorem is given. 

\begin{theorem}\label{thm}
Let $p_M$ and $p_R$ denote the $p$-value of the hypothesis test $H_0$ with samples in MOFA and samples in Random Search, respectively. Then, we have $p_M<p_R$. 
\end{theorem}

From Theorem \ref{thm}, the confidence level of the inference of MOFA is higher than that of random search.

The Key Point of \textbf{Proof:}
This comparison of two populations is called the two sample Behrens-Fisher problem. In statistics, Welch's t-test, or unequal variances t-test \cite{welch1938significance} is a standard solution to this problem. It defines the statistic $t$ by the Eq.~(\ref{eq:4}),
\begin{equation}\label{eq:4}
    {t\quad =\quad ({\;{\bar {Y}}_{1}-{\bar {Y}}_{2})\bigg/{\sqrt {\;{s_{1}^{2} \over N_{1}}\;+\;{s_{2}^{2} \over N_{2}}}}}}.
\end{equation}
The difficulty of calculating the $p$-value is how to calculate the freedom degree of this $t$ statistic. In MOFA, the factorial importance analysis helps to make this easier, and the sampling strategy helps to improve the confidence level. For details please refer to the supplementary materials.

\section{Empirical Evaluation}

In this section, we comprehensively evaluate MOFA on three different settings. 1) a two-layer Bayesian neural network (BNN); 2) the image classification with ResNet on CIFAR10 and 3) an deep neural networks for EEG-based intention recognition. 

\subsection{Experiment Settings}

\textbf{Bayesian Neural Networks. } Following \cite{falkner2018bohb}, we optimize five hyperparameters of a two-layer BNN: the number of units in layer $1$ and layer $2$, the step length, the length of the burn-in period, and the momentum decay. The initial search intervals for these hyperparameters are set to $[2^4, 2^9$], $[2^4, 2^9]$, $[10^{-6}, 10^{-1}]$, $[0, 0.8]$ and $[0, 1]$ respectively. To ensure uniformity of hyperparameter search space, we perform log transformation \cite{falkner2018bohb} on the first three hyperparameters. We reuse an open-source code\footnote{https://github.com/automl/HpBandSter/tree/icml\_2018} to implement the BNN and the baseline methods. Two different UCI datasets, Boston Housing and Protein Structure described in \cite{hernandez2015probabilistic} are used for evaluation. The BNN is trained with Markov Chain Monte-Carlo (MCMC) sampling and the number of steps for the MCMC sampling is used as the budgets, we report the negative log-likelihood on the validation data as the final performance.

\begin{figure}[t!]
\begin{minipage}[t]{0.5\linewidth}
    \includegraphics[width=\linewidth]{bnn-1.pdf}
\end{minipage}%
    \hfill%
\begin{minipage}[t]{0.5\linewidth}
    \includegraphics[width=\linewidth]{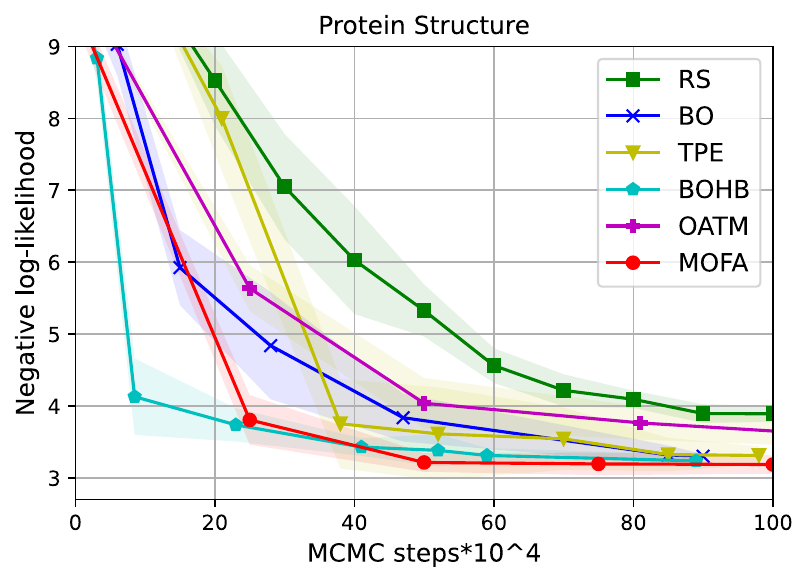}
\end{minipage} 
    \vspace{-1.4em}
    \caption{The negative log-likelihood of BNN on two different UCI datasets ($\lambda=1, l=5, t=2$).}
    \label{fig:bnn}
\end{figure}

\textbf{ResNet on CIFAR-10.} We evaluate MOFA on ResNet (Wide Residual Networks \cite{zagoruyko2016wide} with Cutout \cite{devries2017improved} regularization) for image classification on CIFAR10. The hyperparameters optimized include the base learning rate, the interval of the learning rate, the decay rate of the learning rate, the cutout number and the cutout length. The initial search intervals for these hyperparameters are set to [$10^{-10}$, $10^{-1}$] (with log transformation), [$0.01$, $1$], [$0.01$, $0.99$], [$0$, $3$] and [$1$, $20$] respectively. We set the maximum total budget (epochs) to $12000$. For BOHB, we set the minimum budget to 5 and maximum budget to 135. $5000$ training images are split off as the validation set and the accuracy on the test set is reported for comparison. 


\textbf{EEG-based Intention Recognition.} Following \cite{zhang2019deep}, we evaluate MOFA on a practical task, EEG-based intention recognition with a deep convolutional neural network, the details of the architecture of the networks is described in \cite{zhang2019deep}. A subset of the dataset ($28000$ samples) is used in the experiment. The dimension of each sample is $64$, which corresponds to $64$ channels. The dataset is divided into a training set $(80\%)$ and testing set $(20\%)$, and we report the models' accuracy on the testing set. We optimize three hyperparameters: the learning rate $lr$, the regularization coefficient $\lambda$, and the number of units in each hidden layer. The initial search intervals for the three hyperparameters is set to $[0.0005, 0.01]$, $[0.0005, 0.01]$ and $[64, 1024]$ with log transformation, respectively. 

\textbf{Baselines and Implementation Details.} We compare MOFA with Random Search \cite{bergstra2012random}, BO \cite{mockus1978application}, TPE \cite{bergstra2011algorithms}, BOHB \cite{falkner2018bohb} and OATM \cite{zhang2019deep}. 
For BO and BOHB, the Gaussian Process with Matern kernel function is used as the surrogate model, and expected improvement is used as the aquisition function. 
As OATM cannot be conducted iteratively, we use OAs with different number of runs for fair comparisons. We set the number of runs of OATM to $25$, $50$, $81$, $121$ respectively. 
For MOFA, we build an OLH with index $\lambda=1$, level $l=5$ and strength $t=2$ for comparison. 
In this way, the number of runs in each round will be $N=\lambda l^t=25$. 
In an ablation study (Section~\ref{sec:hpor}) we show how $\lambda$, $l$ and $t$ influence the performance by varying the settings of these numbers. 
All experiments are run $5$ times with different random seeds, and the standard deviations are reported. 

\subsection{Hyperparameter Optimization Results}
\label{sec:hpor}
\begin{figure}[t!]
\begin{minipage}[t]{0.5\linewidth}
\centering
    \includegraphics[width=\linewidth]{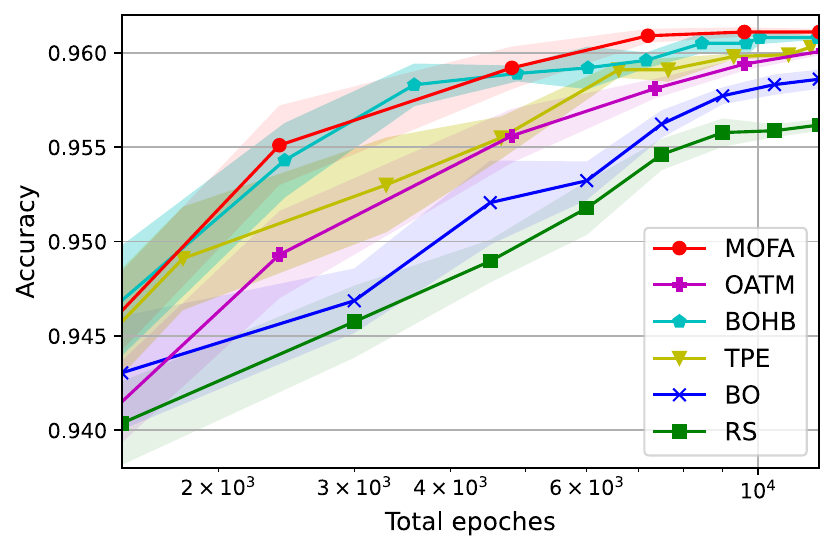}
    \label{fig:resnet}
\end{minipage}%
    \hfill%
\begin{minipage}[t]{0.5\linewidth}
\centering
    \includegraphics[width=\linewidth]{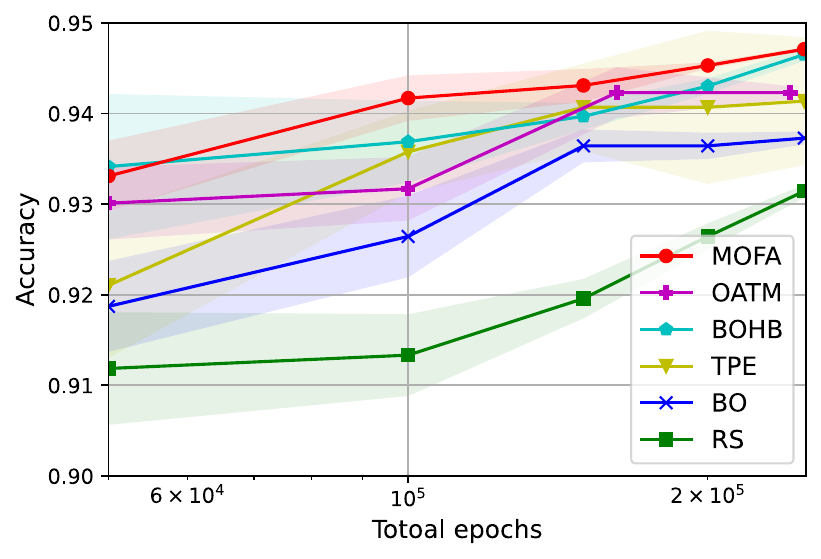}
    \label{fig:cnn-eeg}
\end{minipage} 
 \vspace{-2em}
  \caption{The accuracy of ResNet on CIFAR10 (\textit{left}) and the accuracy of EEG-based intention recognition (\textit{right}). We set the size of OLH to $\lambda=1, l=5, t=2$. }
  \vskip -0.10in
\end{figure}

\textbf{Overall Results. } Fig.~\ref{fig:bnn} shows the HPO results for BNN within two different UCI datasets.  We firstly find that MOFA is significantly better (+12\%) than Random Search with any budget. For the Boston Housing dataset, the performance of MOFA is even 40\% better than that of Random Search when the budget is limited (\textless 60). Compared with BO and TPE, MOFA improves results by +5\% on average, although their performance goes to the same level when increasing budgets. Furthermore, MOFA is even better than BOHB on the Boston Housing dataset and comparable to BOHB on the Protein Structure dataset. 
One should note that BOHB is still model-based and can not be fully parallelized, while MOFA is a highly parallelizable method. 
We also find that OATM outperforms Random Search on two datasets and outperforms BO on Boston Housing but it has no advantage over other model-based algorithms like BOHB, which confirms our assumption that inference analysis (\textit{exploitation}) is important to narrow down the search space. 
Fig.~\ref{fig:resnet} shows the performance for HPO on CIFAR10 with ResNet.The results demonstrate that MOFA consistently outperforms baselines, with only BOHB coming close to MOFA. Fig.~\ref{fig:cnn-eeg} shows the results for HPO on EEG-based intention recognition, showcasing that MOFA outperforms all the baseline solutions including BOHB, improving by +3\% and +2\% compared to RS and BO, respectively.

\begin{figure}[t!]
\begin{minipage}[t]{0.5\linewidth}
    \includegraphics[width=\linewidth]{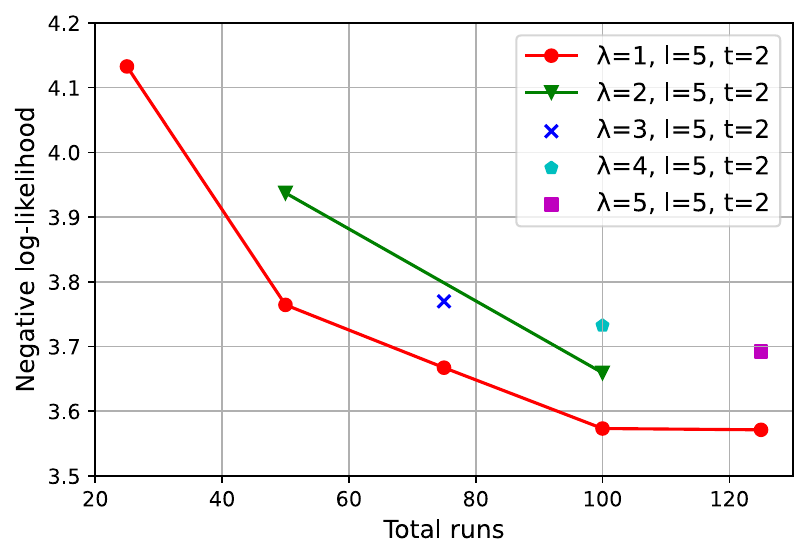}
    \label{fig:olh_ablation}
\end{minipage}%
    \hfill%
\begin{minipage}[t]{0.5\linewidth}
    \includegraphics[width=\linewidth]{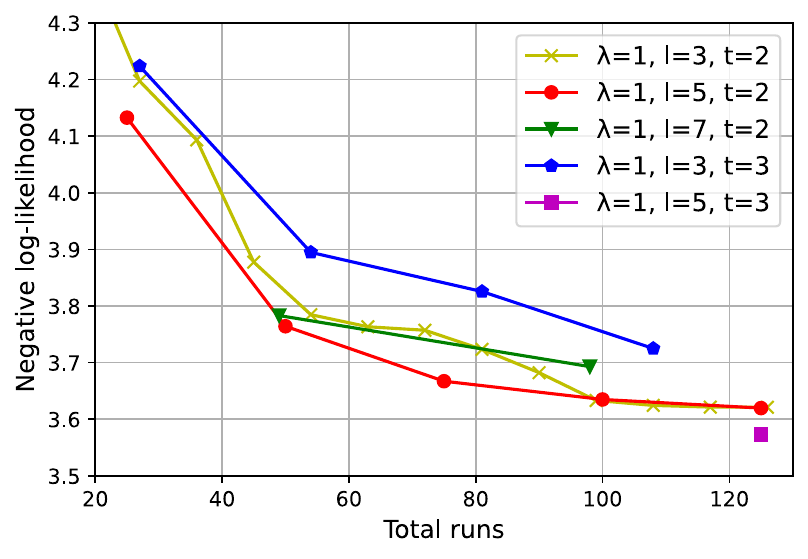}
    \label{fig:olh_ablation_2}
\end{minipage}
\vspace{-2em}
\caption{The negative log-likelihood of MOFA with varying $\lambda$ (left) and MOFA with varying $l$ and $t$ (right) on BNN task. The maximum budgets (number of runs) is fixed to $125$.}
\vskip -0.08in
\end{figure}

\textbf{Ablation Study of Varying Index, Level and Strength.} To evaluate the sensitivity of the model's performance to the index, level and strength. 
We first fix $l=5$ and $t=2$ and change the number of $\lambda$ to $1,2,3,4,5$. Since the number of runs of an OLH is $N=\lambda l^t$, we can get OLHs with runs $25,50,75,100,125$. 
For OLH with $25$ runs ($\lambda=1$), we ran it for five rounds, and for OLH with $50$ ($\lambda=2$) runs, we run it for two rounds, while the OLH with $75$ ($\lambda=3$), $100$ ($\lambda=4$), $125$ ($\lambda=5$) runs are only conducted for one round, which means there is no factorial analysis for these three experiments. 
We further study the the varying number of $l$ and $t$ by fixing $\lambda=1$. For OLH with $t=2$, $l$ is set to $3,5,7$ respectively, while for OLH with $t=3$, $l$ is set to $3,5$. 
Fig.~\ref{fig:olh_ablation} shows the results of MOFA with different numbers of $\lambda$. It clearly shows that OLH ($\lambda=1$) is better than OLH ($\lambda=2$). While OLH ($\lambda=2$) with two rounds outperforms OLH with $\lambda=4$, though taking the same total runs. 
In a nutshell, we find that with the same number of runs, increasing the number of iterations boosts the performance, which also confirms the importance of the factorial analysis (reducing the search space). Fig.~\ref{fig:olh_ablation_2} further shows that OLH with $l=5$ perform better than that with $l=3$ and $l=7$, and the increasing strength of OLH does not boost the performance. 
\par
\begin{figure}[t!]
\begin{minipage}[t!]{0.52\textwidth}
\begin{tabular}{c|ccc}
     \hline
     Method & Boston & Protein & CIFAR-10 \\
     \hline
     RS               &  6.756 & 5.234 & 0.938 \\
     OA-sampler       &  5.133 & 5.125 & 0.944 \\
     LH-sampler       &  5.213 & 5.084 & 0.943 \\
     OLH-sampler      &  4.842 & 5.016 & 0.951 \\
     MOFA (\#1) & 4.232 & 3.824 & 0.955 \\
     MOFA (\#2) & 3.735 & 3.258 & 0.958 \\
     MOFA (\#3) & \textbf{3.658} & \textbf{3.249} & \textbf{0.962} \\
     \hline
\end{tabular}
\captionof{table}{The comparison of MOFA of varying rounds with different sampling schemes.}
\label{tab:mofa_sampler}
\end{minipage}
\begin{minipage}{0.48\textwidth}
\centerline{\includegraphics[width=1.0\columnwidth]{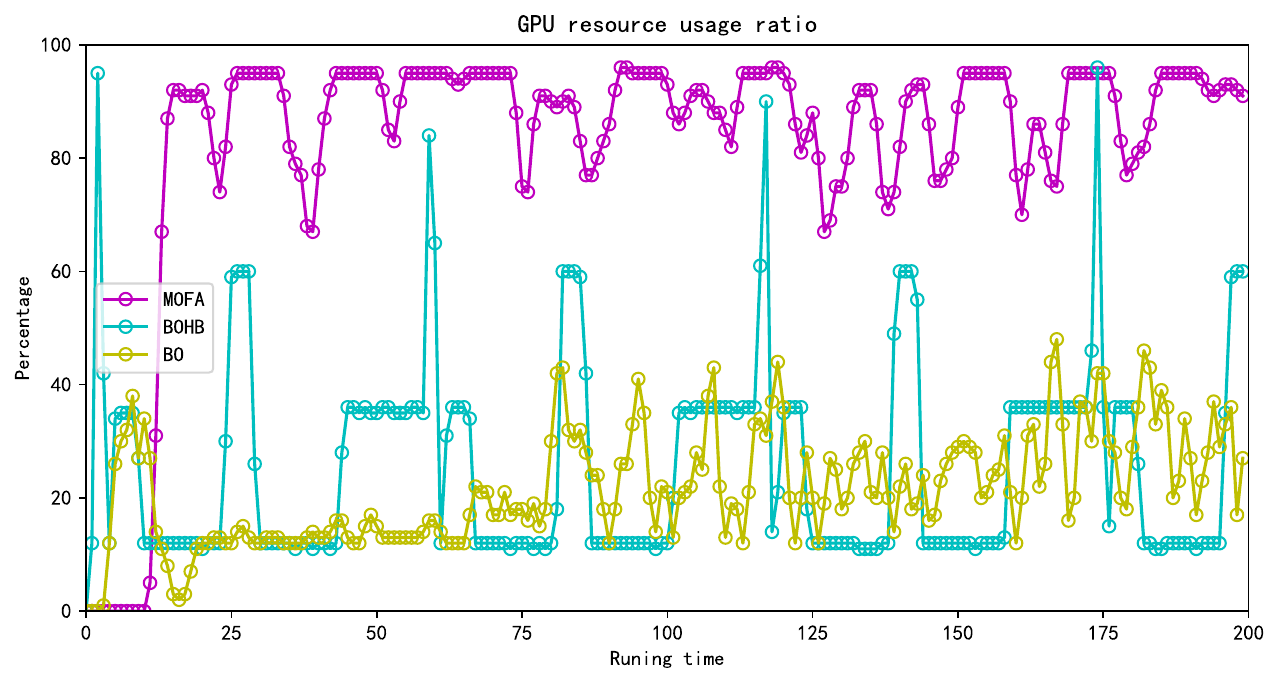}}
\captionof{figure}{The GPU resource usage ratio.}
\label{fig:gpu}
\end{minipage}
\end{figure}

\textbf{Module Contributions Analysis.} To study the module contributions (e.g. how much gain is contributed by factorial design and how much gain is from factorial analysis), we compare MOFA with different sampling schemes including RS, OA-sampler, LH-sampler and OLH-sampler on three datasets (Boston, Protein and CIFAR-10). 
Table.~\ref{tab:mofa_sampler} clearly shows that OLH-sampler performs better than other sampling schemes, demonstrating the benefits of using OLH (factorial design). 
While MOFA outperforms OLH-sampler, showcasing the advantages of using factorial analysis.
\par
\textbf{Computation Efficiency. } In practice, it is very important to take advantage of parallel resources efficiently, as it can significantly reduce the running time of HPO. We compare the GPU resource usage ratio of MOFA with BO (a fully sequential based method) and BOHB (a semi-parallel HPO method). We ran them on the same GPU server with $8$ GPU cards and report the GPU usage ratio. Fig.~\ref{fig:gpu} shows that MOFA almost makes full use of the GPU resources. While for BO, a large proportion of GPU resources are idle. The resource utilization efficiency of BOHB is higher than BO, but it is not as stable as MOFA.
\par
\begin{wrapfigure}{r}{0.48\textwidth}\vspace{-0.5em}
\centering
\caption{Comparison of BO and BO with MOFA initialization for BNN task.}
\includegraphics[scale=0.48]{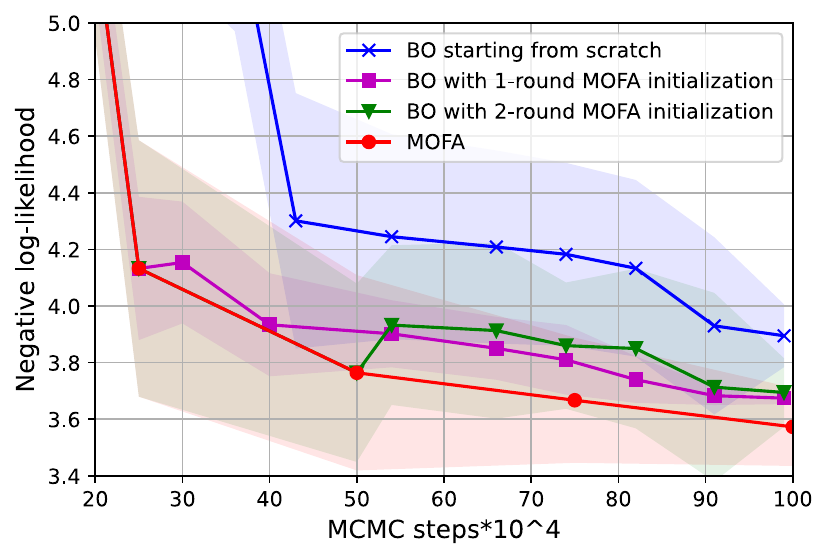}
\vspace{-2.5em}
\label{fig:mofa_bo}
\end{wrapfigure}

\textbf{MOFA as Initialization.} 
Another advantage of MOFA is that it can be easily applied to existing HPO methods as a space initialization module. We apply MOFA as an initialization for BO. Specifically, we firstly conduct MOFA for one or two rounds to narrow down the search space and then conduct BO on the new search space. Fig.~\ref{fig:mofa_bo} shows that with initialization of MOFA, BO outperforms the BO that starts from scratch. Also, we find that using MOFA initialization for one round is much better than using it for two rounds. However, though initialized with MOFA, we find that the performance of BO still cannot exceed the original results of MOFA. One might consider to use MOFA as initialization until the search intervals become small enough, and then continue to search better hyperparameter configurations with other HPO methods.


\subsection{Analysis}
\label{sec: discussions}
\textbf{Model-Free.} Apparently, MOFA follows the standard pipeline of HPO (exploration \& exploitation), but it does not depend on any parametric model. It is lightweight, model-free and easy to implement. 

\textbf{Parallelization.} 
Let the time required by MOFA to prepare one iteration of exploration and to exploit the results be denoted by $a$. Assume that the maximal time needed to evaluate  a single hyperparameter configuration is $t_e$, where $e$ is the $e_{th}$ iteration, the number of processes working in parallel is $n$, the number of iterations is $E$, and the number of runs in OLH is $N$. Then, the maximum total time spent in parallel mode is $T = \sum_{e=1}^E{a+(t_e\times N/n)}$. 
Compared to computing efforts of training and validation, the time required by MOFA (or most other HPO methods) for preparation and analysis $a$ is negligible. Thus, we can approximate $T\approx \sum_{e=1}^E{t_e\times N/n}$. Considering Figures~2 to~4 and Fig.~\ref{fig:gpu} together, one can recognize that increased parallelism of MOFA directly translates into efficiency gains with a speed-up in the order of  $n$. 
\par
\textbf{Sample Efficiency.} Theoretically, the discrepancy of data points can directly affect efficiency of sampling. MOFA uses an OLH-sampler that ensures better discrepancy than other sampling methods, orthogonality of OLH can further improve sample efficiency by eliminating the influence of correlated hyperparameters. Table~\ref{tab:mofa_sampler} shows that the OLH-sampler performs better than other sampling schemes. 

\textbf{Categorical hyperparameters.} MOFA supports both continuous and discrete hyperparameters. In order to support categorical hyperparameters (e.g. selection of optimizers), one solution is to treat the categorical hyperparameters as one-hot hyperparameters. e.g., we treat the \textit{optimizer} as multiple hyperparameters such as \textit{SGD}, \textit{Adam} and \textit{RMSprop}, while each of them has binary ($0,1$) levels.

\section{Related Work}
\textbf{Model-Based Methods.} 
Model-based HPO learns a parametric model such as a surrogate model to search optimal hyperparameters guided by previous experiences. For example, BO \cite{mockus1978application} is a sequential model-based method that balances exploration and exploitation. It is efficient but highly depends on the surrogate model that models the objective function and does not allow for parallelization, since it is a sequential learning-based process. Based on BO, optimization with an early-stopping policy such as Successive Halving (SH) \cite{kumar2018parallel}, HyperBand \cite{li2017hyperband}, BOHB \cite{falkner2018bohb} and ASHA \cite{bergstra2011algorithms} were proposed and achieved significant improvements. Different from BO, TPE \cite{bergstra2011algorithms} learns a surrogate model with a graphic structure, which is more suitable in dealing with conditional variables.
\par
\textbf{Model-Free Methods.}
Model-free HPO tries to tune hyperparameters without any parametric model. Grid Search conducts an exhaustive search on all candidate hyperparameter configurations. It is the most straightforward method but the consumption of computational resources of Grid Search grows exponentially with the increasing number of hyperparameters. Random Search \cite{bergstra2011algorithms} randomly samples a subset of hyperparameter configurations and can assign different budgets to different configurations. However, the randomly selected configurations cannot guarantee to find the global optimum. More recent literature \cite{brockhoff2015impact,konen2011tuned} have applied factorial designs into HPO tasks. \cite{cauwet2020fully} proposed a fully parallel HPO method based on Latin Hypercube, and provided some theoretical analyses, but they only used it to explore the hyperparameter space without corresponding factorial analyses. For factorial analyses, \cite{zhang2019deep} proposed OATM based on OAs, and conducted orthogonal marginal analysis to select best levels for hyperparameters. However, OATM only supports discrete search space and cannot improve hyperparameter configurations by successive iterations. 

\section{Conclusion and Future Work} 
This paper presents a novel HPO method, MOFA, which combines the advantages of being model-free, parallelizable and sample efficient. 
To the best of our knowledge, MOFA is the very first step to exploit factorial design and analysis on HPO, and more sophisticated techniques in these areas can be explored in the future, e.g., exploring the \emph{trade-off} issues (strength $t$ and index $\lambda$) in OLH designs and exploiting more flexible factorial design such as near-orthogonal Latin Hypercubes \cite{cioppa2007efficient}.

\section*{Broader Impact}
Hyperparameter Optimization (HPO) is a pivotal task in automated machine learning. Most of current HPO methods rely heavily on model parametrization and suffer from poor computation efficiency. Rather than learning a parametric sequential-based model, MOFA try to improve the sampling by constructing a low-discrepancy and weak-correlated set of hyperparameters and determining the most promising hyperparameters. MOFA provide a more lightweight solution to HPO, which is model-free and has high efficiency. 

One potential issue of MOFA, like many other HPO methods, is that it requires some human knowledge to determine the best size of OLH such like index, level and strength. We advocate peer researchers to look into this to enhance the intelligence of HPO by reducing the human intervention. 

\section*{Acknowledgments}
The authors thank the International Max Planck Research School for Intelligent Systems (IMPRS-IS) for supporting Bo Xiong. This project has received funding from the European Union’s Horizon 2020 research and innovation programme under the Marie Skłodowska-Curie grant agreement No: 860801.

\newpage

\appendix
\section*{Appendix}

\section{Proof of Theorem \ref{thm}}
The Welch's t-test is used only when the two population variances are not assumed to be equal and hence must be estimated separately. The t statistic to test whether the population means are different is calculated as:
$${ t\quad =\quad ({\;{\bar {Y}}_{1}-{\bar {Y}}_{2})\bigg/{\sqrt {\;{s_{1}^{2} \over N_{1}}\;+\;{s_{2}^{2} \over N_{2}}}}}},$$
where $s_i^2\ (i=1,2)$ is the unbiased estimator of the variance of each of the two samples, and is assumed that $s_1>s_2$ without loss of generality.
This statistic can be also used for testing whether one population mean is larger than the other one. We just need to change the two-sided hypothesis test into the one-sided hypothesis test.

In \cite{owen1995randomly}, it is proved that $$\text{Var}(\bar Y_{M})=\text{Var}(\bar Y_{R})-c/N+o(N^{-1}),$$
where $c$ is a non-negative constant. 
Hence, the t statistic $t_M>t_R$ which means $p_M<p_R$ with the same degree of freedom.
Note that when the value of t statistic is the same, the p-value $p_M<p_R$ is equivalent to the degree of freedom of MOFA is larger than that of Random sampling. Next, we calculate their d.f.s to check this fact.

The degree of freedom is calculated by the Welch–Satterthwaite equation as follows,
$$
{\displaystyle \mathrm {d.f.} ={\frac {\left({\frac {s_{1}^{2}}{N_{1}}}+{\frac {s_{2}^{2}}{N_{2}}}\right)^{2}}{{\frac {\left(s_{1}^{2}/N_{1}\right)^{2}}{N_{1}-1}}+{\frac {\left(s_{2}^{2}/N_{2}\right)^{2}}{N_{2}-1}}}}.}$$
In our cases, the number of trials in different levels $N_1=N_2=N$. 
Then, we have
$$
\mathrm {d.f.}=\frac{(N-1)\left({s_{1}^{2}+s_{2}^{2}}\right)^{2}}{\left(s_{1}^{2}\right)^{2}+ \left(s_{2}^{2}\right)^{2}}.$$

We prove the fact that $\mathrm{d.f.}(M)<\mathrm{d.f.}(R)$ through checking the monotonicity of $
\left({s_{1}^{2}+s_{2}^{2}-2c}\right)^{2}\bigg/\left(\left(s_{1}^{2}-c\right)^{2}+ \left(s_{2}^{2}-c\right)^{2}\right)$ w.r.t the constant $c$ denoted by $d(c)$. To check this, we simplify it into
\begin{align*}
    d(c)&=\frac{\left(s_{1}^{2}-c\right)^{2}+ \left(s_{2}^{2}-c\right)^{2}+2(s_1^2-c)(s_2^2-c)}{\left(s_{1}^{2}-c\right)^{2}+ \left(s_{2}^{2}-c\right)^{2}}\\
    &=1+2\left(\frac{s_{1}^{2}-c}{s_2^2-c}+\frac{s_2^2-c}{s_{1}^{2}-c}\right)^{-1},
\end{align*}
which can be seen that it is monotonically decreasing w.r.t $c$.

Next, the p-value is calculated through the cumulative distribution function of $t$ distribution,
\begin{align*}
p&=\frac{\int_0^{\frac{d.f.}{t^2+d.f.}} x^{d.f./2-1}(1-x)^{-1/2}dx}{2\int_0^1 x^{d.f./2-1}(1-x)^{-1/2}dx}\\
&\triangleq\frac{\int_0^{\frac{d.f.}{t^2+d.f.}} g(x)dx}{2\int_0^1 g(x)dx}.
\end{align*}
Thus, the p-value can be viewed as the area ratio of $g(x)$. For illustration, let $g(x)=b*x$. Then, no matter how $b$ changes, the triangle area ratio is fixed. Note that in our case, $g(x)$ is a convex function and monotonically decreasing w.r.t the degree of freedom. Since $d.f.>2$, we know $g(0)=0$ and $g(1)=\infty$. Hence, the right area of the line $x=\frac{d.f.}{t^2+d.f.}$ plays a major role in the area ratio. According to the monotonicity, when $c$ comes larger, the line $x=\frac{d.f.}{t^2+d.f.}$ moves to the left. Consequently, the area ratio becomes smaller.\hfill $\square$

\end{document}